\pdfoutput=1
\documentclass[11pt]{article}

\usepackage{acl}

\usepackage{times}
\usepackage{latexsym}

\usepackage[T1]{fontenc}
\usepackage[utf8]{inputenc}

\usepackage{microtype}
\usepackage{hyperref}       % hyperlinks
\usepackage{url}            % simple URL typesetting
\usepackage{booktabs}       % professional-quality tables
\usepackage{amsfonts}       % blackboard math symbols
\usepackage{nicefrac}       % compact symbols for 1/2, etc.
\usepackage{microtype}      % microtypography
\usepackage{subfig}
\usepackage{amsmath}
\usepackage{multirow}
\usepackage{soul}
\usepackage{svg}
\usepackage{amsmath}
\usepackage{amssymb}
\usepackage{amsthm}
\usepackage{adjustbox}
\usepackage{bm}
\usepackage{tikz}
\usepackage{lettrine}
\usepackage[titletoc]{appendix}
\usepackage[font=small]{caption}
\usepackage{arydshln}
\usepackage{bbm}
\usepackage{titlesec}
\usepackage{soul}
\usepackage{color}
\usepackage{wrapfig}
\usepackage{graphicx}

\usepackage{pifont}% http://ctan.org/pkg/pifont
\newcommand{\eat}[1]{}
\newcommand{\nop}[1]{}

\title{Modeling Multi-hop Question Answering as Single Sequence Prediction}

\author{
 Semih Yavuz \quad  Kazuma Hashimoto \quad Yingbo Zhou \quad \\
 \textbf{Nitish Shirish Keskar} \quad \textbf{Caiming Xiong} \\
 Salesforce Research \\
 {\texttt {\{syavuz,k.hashimoto,yingbo.zhou,nkeskar,cxiong\}@salesforce.com}}
}

\begin{document}
\maketitle
    \begin{abstract}
Fusion-in-decoder (\textsc{Fid})~\citep{izacard2020fid} is a generative question answering (QA) model that leverages passage retrieval with a pre-trained transformer and pushed the state of the art on single-hop QA.
However, the complexity of multi-hop QA hinders the effectiveness of the generative QA approach.
In this work, we propose a simple generative approach (\textsc{PathFid}) that extends the task beyond just answer generation by explicitly modeling the reasoning process to resolve the answer for multi-hop questions.
By linearizing the \textit{hierarchical reasoning path} of supporting passages, their key sentences, and finally the factoid answer, we cast the problem as a single sequence prediction task.
To facilitate complex reasoning with multiple clues, we further extend the unified flat representation of multiple input documents by encoding cross-passage interactions.
Our extensive experiments demonstrate that \textsc{PathFid} leads to strong performance gains on two multi-hop QA datasets: HotpotQA and IIRC.
Besides the performance gains, \textsc{PathFid} is more interpretable, which in turn yields answers that are more faithfully grounded to the supporting passages and facts compared to the baseline \textsc{Fid} model.
\end{abstract}

%% FOR SUBMISSION -- Copy then FIX
\iffalse
Fusion-in-decoder (Fid) (Izacard and Grave, 2020) is a generative question answering (QA) model that leverages passage retrieval with a pre-trained transformer and pushed the state of the art on single-hop QA. However, the complexity of multi-hop QA hinders the effectiveness of the generative QA approach. In this work, we propose a simple generative approach (PathFid) that extends the task beyond just answer generation by explicitly modeling the reasoning process to resolve the answer for multi-hop questions. By linearizing the hierarchical reasoning path of supporting passages, their key sentences, and finally the factoid answer, we cast the problem as a single sequence prediction task. To facilitate complex reasoning with multiple clues, we further extend the unified flat representation of multiple input documents by encoding cross-passage interactions. Our extensive experiments demonstrate that PathFid leads to strong performance gains on two multi-hop QA datasets: HotpotQA and IIRC. Besides the performance gains, PathFid is more interpretable, which in turn yields answers that are more faithfully grounded to the supporting passages and facts compared to the baseline Fid model.
\fi

    \section{Introduction}
Leveraging knowledge to make complex reasoning has been a fundamental problem of artificial intelligence.
Open-domain question answering (QA) \citep{voorhees1999trec8} is an integral part of such a line of research with impactful applications~\citep{co-search,co-index}, where the task is to answer general domain questions by gathering evidence from a large collection of documents.
While super-human level performance has been achieved on single-passage reading comprehension dataset like SQuAD~\citep{rajpurkar-etal-2016-squad}, open-domain QA still has a long way to go, especially for questions requiring more complex reasoning.
The main challenge in the task of complex QA, namely {\it multi-hop QA}, is that it requires a QA system to combine multiple pieces of evidence from multiple documents~\citep{welbl-etal-2018-constructing, talmor-berant-2018-web, yang-etal-2018-hotpotqa}.
Even for single-hop QA, it has been shown challenging for extractive QA models to effectively aggregate evidence from the combined pool of multiple passages, which has been the focus of recent work \citep{clark-gardner-2018-simple, min-etal-2019-discrete, guu2020realm}.

\begin{figure*}[t]
    \center
	\includegraphics[width=\textwidth]{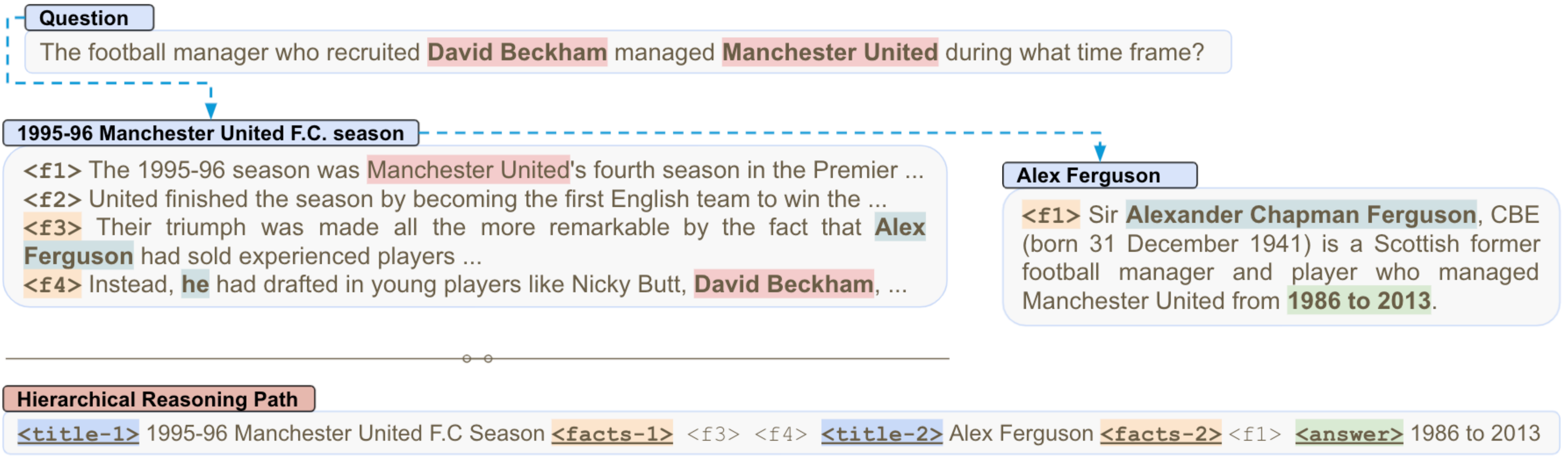}
	\caption{An example of multi-hop question from HotpotQA dataset. It requires fusing multiple evidences (supporting facts) from multiple passages in a certain order to arrive at the correct answer. We formulate the entire problem as a single sequence prediction of the linearized hierarchical path ending with the answer.}
	\label{figure:hotpotqa_example}
\end{figure*}

Recent work~\citep{lewis2020rag,min-etal-2020-ambigqa} has demonstrated the promise of a generative approach at combining evidences from multiple passages for answer generation.
Thanks to large pre-trained transformers like T5~\citep{raffel2020t5}, \citet{izacard2020fid} introduced {\it fusion-in-decoder} (\textsc{Fid}) that leverages passage retrieval with generative models for open-domain QA, achieving state-of-the-art scores across several single-hop QA benchmarks.
However, we observe that the success of the \textsc{Fid} model does not extend to multi-hop QA, which is corroborated by the findings in \cite{xiong2021mdr}.
Further, the \textsc{Fid} model is a rather opaque model in terms of interpretation of the answer generation process.
This capability becomes especially important for multi-hop QA, which requires sequential reasoning across multiple evidences from the pool of retrieved passages.

In this work, we propose \textsc{PathFid}, a generative QA model that learns to generate an answer along with a {\it reasoning path} to improve its capability of multi-hop reasoning.
\textsc{PathFid} extends multi-hop QA beyond just answer generation by explicitly modeling the full reasoning path to resolve the answer with a generative sequence-to-sequence model.
To this end, we cast the problem as a single sequence prediction task that simultaneously models reasoning path consisting of supporting passages and facts, and eventually the factoid answer.
Furthermore, we extend \textsc{PathFid} to allow for cross-passage interactions between the retrieved passages to obtain more expressive representations from the encoder to facilitate modeling a complex reasoning chain by the decoder.
Figure~\ref{figure:hotpotqa_example} shows an example of our task formulation, and Figure~\ref{figure:model_overview} shows an overview of our approach.
We evaluate our proposed approach on two multi-hop QA datasets: HotpotQA~\citep{yang-etal-2018-hotpotqa} and IIRC~\citep{ferguson-2020-iirc}.
Our extensive experiments demonstrate that (i) \textsc{PathFid} leads to significant performance gains over \textsc{Fid} on answer generation, (ii) \textsc{PathFid} is the first generative model unlocking the possibility of generating the reasoning path jointly with the answer while achieving competitive performance on supporting fact extraction metric as well.
Besides the performance gains, \textsc{PathFid} is able to expose the underlying reasoning process behind the answer generation, which allows us to conduct a much finer-grained qualitative and quantitative analysis on the model's behavior, providing insights into further improving and better understanding generative models for multi-hop QA.

    \section{Problem Setup and Background}

In this section, we formally introduce the problem setup and establish the necessary background.

\subsection{Multi-hop Question Answering}
We first describe the multi-hop QA task in a general way.
We assume that a collection of $K$ passages are given for a question $q$: $D_q = \{p_1, p_2, \ldots, p_K\}$,
where $D_q$ can be a pre-defined set, or it can also be an output from a text retrieval system (e.g., DPR~\citep{karpukhin-etal-2020-dense} and MDR~\citep{xiong2021mdr}) in an open-domain QA setting.
That is, in the case of the open-domain setting, $D_q$ is a subset of a large collection of passages, such as Wikipedia.
The task is to generate an answer string $a$ given $q$ and $D_q$.
In addition, we aim at identifying which passages provide evidence, and which sentences in them are describing the evidence.
Figure~\ref{figure:hotpotqa_example} shows a comprehensive example of the task definition, where we can see that some sentences (called {\it supporting facts}) in the two paragraphs are crucial to answer the question.
Moreover, there is a reasoning flow: the question $\rightarrow$ the first paragraph $\rightarrow$ the second paragraph, which is called a {\it reasoning path} in previous work~\citep{asai2020grr}.
The overall task is then to predict the reasoning path along with the supporting facts, and the answer.

\subsection{Fusion-in-Decoder Model (\textsc{Fid})} \label{subsection:fid}
Fusion-in-Decoder (\textsc{Fid}) is a generative reader based on a sequence-to-sequence architecture, initialized from pre-trained models such as T5 \citep{raffel2020t5} or BART \citep{lewis-etal-2020-bart}.
It consists of an encoder ($\mathbf{Enc}$) and a decoder ($\mathbf{Dec}$).
First, it constructs a single block of text 
$b_n:=\texttt{question:} \ q \ \texttt{title:} \ t_n \ \texttt{context:} \ p_n$ 
of concatenated evidence from each passage-title pair $(p_n, t_n)$ together with the question ($q$).
Then, each of the resulting evidence block $b_n$ is independently encoded into $|b_n| \times d$-dimensional output representations, which are then concatenated to form a unified input representation 
\begin{align} \label{eqn:fid-X}
    \mathbf{X} = [\mathbf{Enc}(b_1); \mathbf{Enc}(b_2); \ldots, \mathbf{Enc}(b_N)] 
\end{align}
of dimension $(\sum_{n}|b_n|) \times d$ 
where $|b_n|$ denotes the length of the $n$-th block $b_n$ in number of tokens.
Note that, the motivation behind this strategy is to avoid the expensive quadratic self-attention computation on the encoder-side, effectively reducing the complexity from $\mathcal{O}(\left(\sum{|b_n|}\right)^2)$ to $\mathcal{O}(\sum{|b_n|}^2)$.
Then, the overall answer generation is modeled as a conditional generation $p_{\theta}(a | \mathbf{X})$ given $\mathbf{X}$ consuming the unified input representation $\mathbf{X}$, where $\theta$ represents the set of all model parameters.
The model is trained to minimize the cross-entropy loss for generating answer tokens on the decoder side. 
At inference time, \textsc{Fid} first computes $\mathbf{X}$ based on the retrieved passages, and then decodes the answer token by token following $p_{\theta}(a_i | a_{<i}, \mathbf{X})$ with the learned model parameters $\theta$.

    \section{\textsc{PathFid} Reader for Multi-hop QA} \label{section:approach_pathfid}
In this section, we introduce a generative reader (\textsc{PathFid}) for $K$-hop QA that jointly generates an alternating sequence of \textit{passage-level} and \textit{fact-level} clues on the reasoning path by more explicit fusion of evidence from the pool of input passages to arrive at the correct answer.

\subsection{Overview of \textsc{PathFid}}
\begin{figure*}[t]
    \center
	\includegraphics[width=\textwidth]{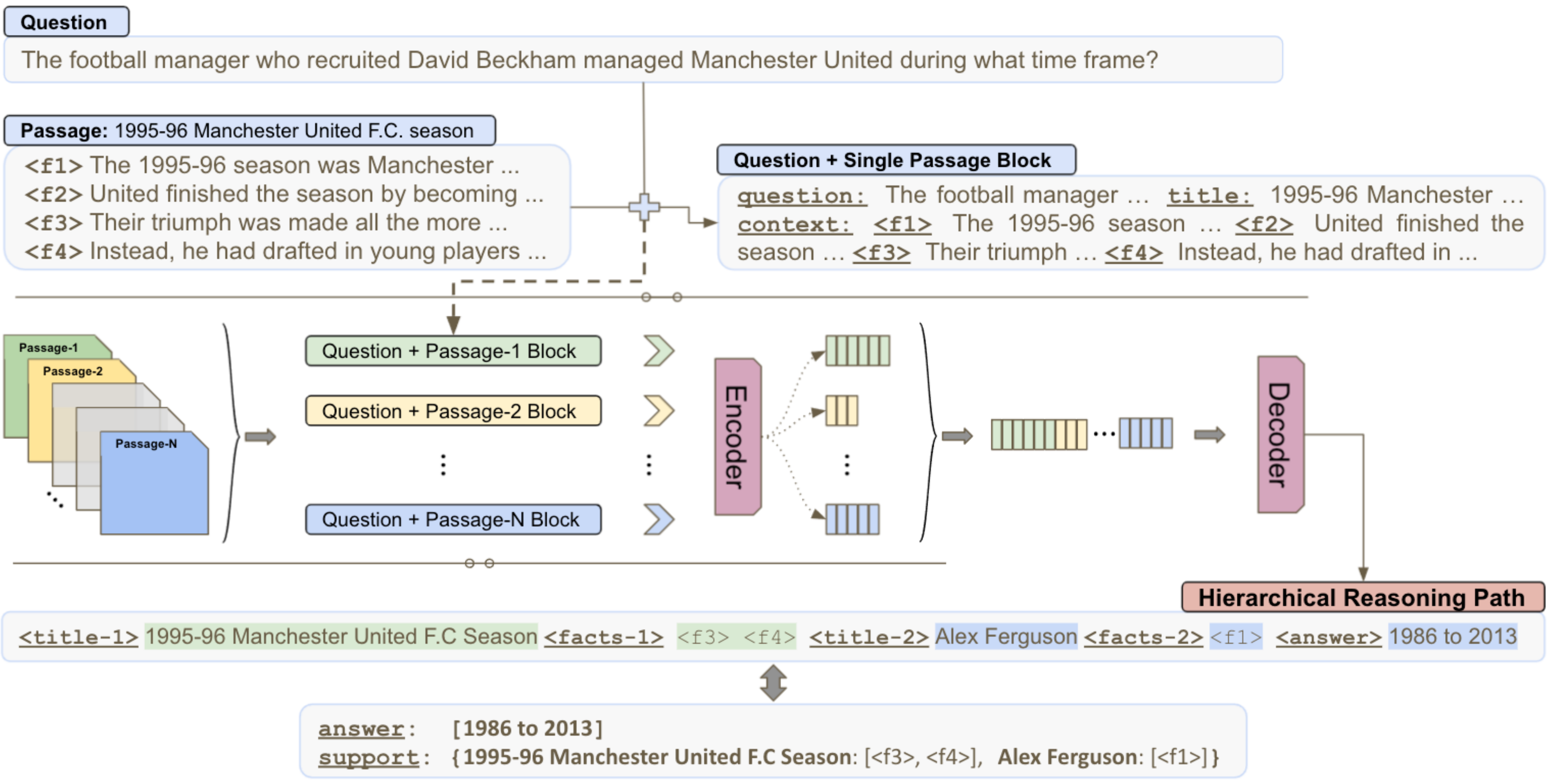}
% 	\vspace{-3mm}
	\caption{\textsc{PathFid} model overview. Each question+passage block is encoded in parallel, which are then concatenated in to a long flat sequence of vector representations. The decoder then consumes this long sequence and generates the full reasoning path, which is then uniquely parsed into the final answer along with the supporting facts exposing the underlying reasoning.} 
	\label{figure:model_overview}
% 	\vspace{-3mm}
\end{figure*}
As illustrated in Figure \ref{figure:model_overview}, \textsc{PathFid} employs a single sequence-to-sequence architecture that independently encodes the input passages after inserting special fact markers (\texttt{<}$f_i$\texttt{>}) before the $i$-th sentence of each passage.
Conditioning on the concatenation of token-level input representations per passage, its decoder then generates the linearized hierarchical reasoning path obtained by concatenating the sequence of passage titles and their corresponding supporting fact pointers followed by the answer.
Each segment on the reasoning path is separated by special markers in a way that makes it possible to uniquely recover the individual segment predictions after decoding in the inference time.

\subsection{Extending Multi-hop QA beyond Answer Generation}
The opaqueness of the \textsc{Fid} model, which makes understanding of the reasoning process more difficult, motivated our approach and its emphasis on exposing the reasoning path.
Instead of only modeling answer generation, we propose to jointly model it with the full \textit{reasoning path} in an \textit{hierarchical fashion} to derive the answer in a unified way using \textit{multi-task maximum likelihood training}.

\subsubsection{Global Input Representation}
We utilize the core input encoding architecture from \textsc{Fid} approach (Section \ref{subsection:fid}) by introducing a new passage representation that will facilitate supporting fact generation on the reasoning path as illustrated in Figure \ref{figure:model_overview}.
To this end, we independently encode each input passage-title pair $(p_n, t_n)$ along with the question $q$ as a separate block 
% \begin{align} \label{eqn:path_block}
%     b^{\text{path}}_n := \texttt{question:} \ q \ \texttt{title:} \ t_n \ \texttt{context:} \ p^{\text{path}}_n
% \end{align}
$b^{\text{path}}_n := \texttt{question:} \ q \ \texttt{title:} \ t_n \ \texttt{context:} \ p^{\text{path}}_n$
% \nitish{this is a weird notation. can we improve somehow?}
where we redefine the context representation by inserting special tokens (\texttt{<}$f_i$\texttt{>}) before each sentence of the passage as
\begin{align} \label{eqn:context_in_path_block}
    \hspace{-2mm} p^{\text{path}}_n := \ \texttt{<}f_1\texttt{>} \ s_{n}^{(1)} \ \texttt{<}f_2\texttt{>} \ s_{n}^{(2)} \ \cdots \ \texttt{<}f_{l_n}\texttt{>} \ s_{n}^{(l_n)}
\end{align}
where $s_{n}^{(i)}$ denotes the $i$-th sentence of passage $p_n$, and $l_n$ is the number sentences it contains. 
Having redefined the input blocks ($b^{\text{path}}_n$) per passage, we then compute the global input representation similar to Eq. \ref{eqn:fid-X} by 
\begin{align} \label{eqn:path-fid-X}
    \mathbf{X}_{q}^{\text{path}} = [\mathbf{Enc}(b_1^{\text{path}}); \mathbf{Enc}(b_2^{\text{path}}); \ldots; \mathbf{Enc}(b_N^{\text{path}})] 
\end{align}
Note that sentence indicators (\texttt{<}$f_i$\texttt{>}) are shared across all passages, encouraging a more hierarchical passage representation by explicitly breaking them down into sentence-level sub-blocks using the same indicator tokens.

\subsubsection{Hierarchical Reasoning Path as a Sequence}
The hierarchical design of reasoning path is inspired by the human reasoning process for multi-hop QA task.
More precisely, if a question $q$ requires $K$-hop reasoning, then we process these $K$ passages in a sequential order alternating between their passage-level and sentence-level evidence until we reach the answer.
To this end, let $R_q = \{p_{r_1}, p_{r_2}, \ldots, p_{r_K}\}$ with $r_i \in [1, N]$ denote the sequence of passages from the larger pool $D_q$ reflecting this reasoning process for locating the answer $a$ for question $q$.
As shown in Figure \ref{figure:model_overview}, we define the hierarchical reasoning path as a linearized sequence of blocks of passage titles and supporting facts followed by the answer block
\begin{align}
    \hspace{-3mm}\mathbf{Y}_{q}^{\text{path}} := [T_{r_1}; E_{r_1}; T_{r_2}; E_{r_2}; \cdots; T_{K}; E_{r_K};  A]
\end{align}
where $T_{r_i}$ represents the $i$-th title block obtained by inserting a special token (\texttt{<title-i>}) before the title $t_{r_j}$ and $A$ denotes the answer block derived by prepending a special token (\texttt{<answer>}) to the answer $a$ as illustrated in Figure \ref{figure:model_overview}.
On the other hand, $i$-th supporting fact block is defined as the sequence of fact indicators following \texttt{<facts-i>} token by
\begin{align} \label{eqn:supporting_fact_block}
    \hspace{-3mm}E_{r_i} := \texttt{<facts-i>} \ \texttt{<}f_{j_1}\texttt{>} \ \texttt{<}f_{j_2}\texttt{>} \cdots \ \texttt{<}f_{j_{m_i}}\texttt{>}
\end{align}
where $\{j_1, j_2, \ldots, j_{m_i}\}$ denote the indices of key sentences to leverage from passage $p_{r_i}$ to transition to the next evidence on the reasoning process $R_q$ for question $q$, and $1 \leq m_i \leq l_{r_i}$ denotes the number of supporting facts.
Note that fact indicators $\texttt{<}f_{i}\texttt{>}$ are shared between the contexts $p^{\text{path}}_n$ of input blocks (Eq. \ref{eqn:context_in_path_block}) and supporting fact blocks (Eq. \ref{eqn:supporting_fact_block}) on the target reasoning path to allow the decoder to follow along the sequential reasoning $R_q$ by pointing to the facts $E_{r_i}$ of passage $p_{r_i}$.

\subsection{Encoding Cross-Passage Interactions (\textsc{PathFid+})} \label{subsection:path_fid_plus}
\textsc{PathFid} enables more explicit evidence fusion through the reasoning path to guide the model to towards correct answer in a structured way.
However, it still relies on the decoder to combine all the clues together, which might still struggle due to lack of cross-passage interactions as input blocks are encoded independently.
To address this potential limitation, we propose \textsc{PathFid+}, where we further extend \textsc{PathFid} in a way that enables cross-passage interaction by redefining the input block consisting of a pair of passages $(p_{n_1}, p_{n_2})$ as
\begin{align*} % \label{eqn:path_plus_block}
    b^{\text{path+}}_{n_1, n_2} :=& \ \texttt{question:} \ q \\
                                &\texttt{<title-1>} \ t_{n_1} \ \texttt{<context-1>} \ p^{\text{path}}_{n_1} \\
                                &\texttt{<title-2>} \ t_{n_2} \ \texttt{<context-2>} \ p^{\text{path}}_{n_2}
\end{align*}
assuming that a set of passage pairs $(p_{n_1}, p_{n_2})$ are available for model to consume.
In particular, we derive a set of pairs of passages from the initial set $D_q$ by $D_q^{+} = \{(p^{*}, p_1), (p^{*}, p_2), \ldots, (p^{*}, p_N)\}$
where $p^{*}$ corresponds to the first passage that is possible to immediately hop to from question $q$, which may be determined by another model, or by executing the original \textsc{PathFid} on $D_q$ in our case.
Global input representation $\mathbf{X}_{q}^{\text{path+}}$ is obtained similarly (Eq. \ref{eqn:path-fid-X}) by except encoding the new blocks $b^{\text{path+}}_{n_1, n_2}$ allowing for cross-passage interactions, while the target reasoning path $\mathbf{Y}_{q}^{\text{path+}}$ remains the same as $\mathbf{Y}_{q}^{\text{path}}$.
Note that \texttt{<title-i>} special markers are shared between new input block $b^{\text{path+}}_{n_1, n_2}$ and target reasoning path $\mathbf{Y}_{q}^{\text{path+}}$ to provide the model with additional clue regarding the first passage on the reasoning path while still relaying the complete evidence fusion to the decoder via information redundancy encoded in $\mathbf{X}_{q}^{\text{path+}}$.

\subsection{Training and Inference}
Having defined global input representation $\mathbf{X}_{q}^{\text{path}}$, the decoder autoregressively generates the reasoning path $\mathbf{Y}_{q}^{\text{path}}$ per token at each step by following self-attention, cross-attention on the entire $\mathbf{X}_{q}^{\text{path}}$, and feed-forward modules. 
So, the overall reasoning path generation is modeled as conditional generation $p_{\theta^{\text{path}}}(\mathbf{Y}_{q}^{\text{path}} | \mathbf{X}_{q}^{\text{path}})$.
The model then is trained to minimize 
$J(\theta^{\text{path}}) = -\sum_{i=1}^{|\mathbf{Y}_{q}^{\text{path}}|} \log p_{\theta}(y_i | y_{<i}, \mathbf{X}_{q}^{\text{path}})$
with teacher forcing over a training set of $\{(q, a, D_q)\}$.

In the inference, the decoder consumes the input representation $\mathbf{X}_{q}^{\text{path}}$ computed by encoder, and generates the full reasoning path token by token. 
We then post-process the decoded sequence using the answer indicator (\texttt{<answer>}) to first obtain the answer, followed by recursively parsing the remaining sequence using the special separator tokens (\texttt{<title-k>}, \texttt{<facts-k>}) to reconstruct the title and retrieve its relevant sentences at each hop $k$.
As illustrated in Figure \ref{figure:model_overview}, the final result of the inference can be summarized into a dictionary which maps each generated passage title to the list of sentence pointers as well as the final answer.

    \section{Experiments}
\subsection{Datasets and General Setup} \label{subsection:general_setup}
We conduct experiments on two multi-hop question answering datasets: \textit{HotpotQA} and \textit{IIRC}.

\noindent{\textbf{HotpotQA}} \citep{yang-etal-2018-hotpotqa} is a large-scale human-annotated dataset including 113K multi-hop questions. 
It focuses on using documents from Wikipedia as the source of information for answering questions rather than knowledge bases as in other multi-hop QA datasets \citep{welbl-etal-2018-constructing, talmor-berant-2018-web}.
The questions in HotpotQA are not constrained by the fixed knowledge-base schema, hence they can cover more diverse topics.
The answer for each question in HotpotQA is extracted from 10 paragraphs in the \textit{distractor} setting, while it is allowed to use the entire Wikipedia for the \textit{full wiki} setting.
There are two main question types \textit{bridge} (80\%) and \textit{comparison} (20\%) in the corpus, where each question is designed in a way that extracting the correct answer requires reasoning over multiple evidence distributed across two passages.
While comparison questions do not require the these passages to be processed in a particular order, \textit{bridge} questions often require identifying the bridge entity in the first passage to correctly hop to the second one that contains the answer.
Each question is also provided with the annotation of 2 supporting passages and up to 5 corresponding relevant sentences as their supporting facts. 
Since our proposed approach is a reader model that reasons over a given set of evidence documents, we primarily focus our experiments on the \textit{distractor} setting\footnote{See Appendix \ref{appendix:full_wiki} for \textsc{PathFid} results in open-domain setting using MDR \citep{xiong2021mdr} as the retriever.}.

\noindent{\textbf{IIRC}} \citep{ferguson-2020-iirc} is a dataset of more than 13K human-written questions over paragraphs from English Wikipedia, where crowdworkers had access only to initial paragraph and list of hyperlinks to other relevant Wikipedia articles, with the missing information occurring in one or more linked documents. 
This annotation design encouraged less lexical overlap between the questions and the contexts that actually contain the answer. 
This dataset presents unique challenges compared to HotpotQA because (1) it additionally requires discrete/numerical reasoning and identification of unanswerable questions, which adds up to 4 different possible answer types (span, binary, numerical, unanswerable), and (2) about 30\% of questions require reasoning over more than 2 passages including the main passage.

\noindent{\textbf{Evaluation Metrics.}} 
We use standard metrics exact-match (EM) and $\text{F}_1$ scores for measuring the quality of predicted answers.
For HotpotQA experiments, we are also able to evaluate \textsc{PathFid} on supporting fact predictions using the official metrics (Support-EM, Support-$\text{F}_1$), which measures the performance of the reader model in correctly identifying the supporting facts from the relevant passages. 
Note that this metric implicitly requires correctly identifying relevant passages among the distractors as well.
For our experiments on IIRC dataset, similar to the baseline model constructed in the original work \citep{ferguson-2020-iirc}, we follow the evaluation methods used by DROP~\citep{dua-2019-drop}. 

\noindent{\textbf{Implementation Details.}}
We use pre-trained T5-large encoder-decoder \citep{raffel2020t5} to initialize the models in our experiments.
We train the model with batch size of 64 with constant learning rate of 1e-4 for 10 epochs.
We use maximum length of 256 (resp. 512) tokens for input blocks of \textsc{PathFid} (resp. \textsc{PathFid+}), while the maximum target sequence length is set to be 64.
However, the sequence truncation is performed on the reasoning path excluding answer part for sequences of length longer than 64 tokens.
All the experiments are conducted on a machine with 4 or 8 many 40GB A100 GPUs. 
Our code is based on Huggingface Transformers \citep{wolf2020huggingfaces}. 
Please see Appendix for further details on the hyperparameter settings.

\subsection{Main Experiments: HotpotQA}

\subsubsection{Overall Results}
\begin{table*}[!t]
	\centering
	\begin{adjustbox}{max width=\textwidth}
		\begin{tabular}{lcccc}
			\hline  \\ [-2ex]
			\multicolumn{1}{c}{} & \multicolumn{2}{c}{\textbf{Answer}} & \multicolumn{2}{c}{\textbf{Support}} \\
			\cline{2-5} \\ [-2ex]
			\textbf{Methods} &  EM &  F1 & EM &  F1 \\
			\hline
			\hline  \\ [-2ex]
            Baseline \citep{yang-etal-2018-hotpotqa}                              & 44.4 & 58.3 & 22.0 & 66.7  \\
            DFGN \citep{qiu-etal-2019-dynamically}                 & 55.4 & 69.2 & - & -  \\
            QFE \citep{nishida-etal-2019-answering}                & 53.7 & 68.7 & 58.8 & 84.7  \\
            SAE \citep{tu2020sae}                                  & 61.3 & 74.8 & 58.1 & 85.3 \\
            SAE-large \citep{tu2020sae}                            & 67.7 & 80.8 & 63.3 & 87.4 \\
            Graph Recurrent Retriever \citep{asai2020grr} (base)   & 52.7 & 65.8 & 57.4 & 84.6  \\
            Graph Recurrent Retriever \citep{asai2020grr} (wwm)    & 68.0 & 81.2 & 58.6 & 85.2  \\
            Gated Memory Flow \citep{shao2021memory}               & 69.6 & 83.0 & 64.7 & 89.0  \\
			\hline
			\textbf{This Work} \\
			\textsc{Fid}* \citep{izacard2020fid} & 64.4 & 77.8 & - & -  \\
			\textsc{PathFid}  & 65.8 & 78.9 & 59.3 & 85.7  \\
            \textsc{PathFid+}  & 72.7 & 84.2 & 64.9 & 88.7  \\
			\hline
		\end{tabular}
	\end{adjustbox}
	\caption[Table caption text]{Results on the development set of HotpotQA distractor setting in comparison with previous work. \textsc{Fid}* indicates that the reported results are obtained by our implementation following the training details in the paper.}
	\label{table:main_res}
% 	\vspace{-3mm}
\end{table*}

We present our main results on the HotpotQA distractor setting in Table \ref{table:main_res}. 
We report results on the HotpotQA development set in comparison with the previous published methods. 
\textsc{PathFid} reader provides 1.4\% absolute gain on answer EM score in comparison to \textsc{Fid} model. 
Moreover, it achieves competitive supporting fact predictions of 59.3\% support-EM and 85.7\% support-$\text{F}_1$ as a result of path generation compared to strong extractive models such as \citep{asai2020grr}.
In summary, \textsc{PathFid} establishes the usefulness of modeling the full reasoning path along with answer generation for multi-hop QA. 
More notably, \textsc{PathFid+} achieves a quite significant performance gain across all the central evaluation metrics, demonstrating the importance of cross-passage interactions.
Overall results validate the effectiveness of the two central modeling contributions of our proposed method. 
Next, we present further analysis and discussion on the unique advantages of \textsc{PathFid} approach under a few central questions which motivated our research at the first place.

\subsubsection{Analysis}
\noindent{\textbf{How faithfully grounded are the generated answers on supporting facts?}}
\begin{table}[!t]
	\centering
	\begin{adjustbox}{max width=\columnwidth}
		\begin{tabular}{lccc}
			\hline  \\ [-2ex]
			\textbf{Criterion} & \textsc{Fid} & \textsc{PathFid} & \textsc{PathFid+}  \\
			\hline
            Pred Answer Grounded in Gold Passages  & 93.9 & 95.3 & 97.7 \\
            Pred Answer Grounded in Gold Supports  & 90.8 & 92.1 & 95.6 \\
            \hline  \\ [-2ex]
            Gold Answer Grounded in Pred Passages  & - & 96.2 & 98.0 \\
            Gold Answer Grounded in Pred Supports  & - & 95.3 & 97.4 \\
            \hline  \\ [-2ex]
            Pred Answer Grounded in Pred Passages  & - & 96.4 & 97.5 \\
            Pred Answer Grounded in Pred Supports  & - & 90.3 & 94.3 \\
			\hline
		\end{tabular}
	\end{adjustbox}
% 	\vspace{1mm}
	\caption[Table caption text]{How faithfully grounded are the gold/predicted answers in gold/predicted supporting facts?}
	\label{table:groundedness}
\end{table}
\begin{table*}[t]
	\centering
	\begin{adjustbox}{max width=\textwidth}
		\begin{tabular}{lcccccccc}
			\hline  \\ [-2ex]
			\multicolumn{1}{c}{} & \multicolumn{4}{c}{\textbf{Answer-EM}} & \multicolumn{4}{c}{\textbf{Support-EM}} \\
			\cline{3-5} \cline{7-9} \\ [-2ex]
			\multicolumn{1}{c}{} & \multicolumn{2}{c}{\textbf{Comparison}} & \multicolumn{2}{c}{\textbf{Bridge}} & \multicolumn{2}{c}{\textbf{Comparison}} & \multicolumn{2}{c}{\textbf{Bridge}} \\
			\cline{2-9} \\ [-2ex]
			\textbf{\# Supp Facts} &  \textsc{Fid} & \textsc{PathFid} & \textsc{Fid} & \textsc{PathFid} &  \textsc{Fid} & \textsc{PathFid} &  \textsc{Fid} & \textsc{PathFid} \\
			\hline
			\hline  \\ [-2ex]
			2 & 70.4 & 71.8 & 63.3 & 64.6  & - & 86.7  & - & 70.0  \\
			3 & 66.1 & 68.2 & 62.7 & 63.1  & - & 43.4  & - & 30.7  \\
			4 & 62.2 & 63.8  & 64.3 & 66.5  & - & 5.4 & - & 26.2  \\
            >=5 & 83.3 & 87.5 & 60.0 & 65.0  & - & 0.0  & - & 3.8  \\
            \hline
		\end{tabular}
	\end{adjustbox}
	\caption[Table caption text]{Performance breakdown on Answer-EM and Support-EM by question type and the number of gold supporting facts (rows). Since \textsc{Fid} does not generate supporting facts, corresponding columns are left empty.}
	\label{table:breakdown}
% 	\vspace{-2mm}
\end{table*}
In Table \ref{table:groundedness}, we present a detailed analysis comparing different models in terms of the faithfulness of their generated answers on both gold and predicted supporting facts.
The first row focuses on the passage-level answer grounding computed by the percentage of the answers found in one of the gold supporting passages, while the second row reports the same analysis on sentence-level. 
We can observe that \textsc{PathFid} models significantly improves on how faithfully the generated answers are grounded on the supporting facts both at passage-level and sentence-level granularities.
The next two rows provide further insight into the quality of the generated supporting facts by \textsc{PathFid} models by measuring how often the gold answer can be found in them.
This analysis shows that the generated supporting facts are of quite high-quality including the gold answer for more than 95.3\% and 96.2\% at sentence-level and passage-level, respectively.
The last two rows measure the faithfulness of the generated answers on the model generated supporting facts, which is not applicable to \textsc{Fid} model as it does not perform supporting fact prediction. 
We observe that the generated answers are quite faithfully grounded on the predicted supporting facts, showing the path generation not only improves the answer EM performance but also successfully grounds them on the evidence it generates as part of the full reasoning path.

It is important emphasize here that extractive reader models can be guaranteed to output perfectly grounded answers simply by locating the answer in their predicted supporting facts.
On the other hand, it is difficult for generative models to ensure 100\% answer grounding simply due to its generative nature.
However, we are able to provide additional evidence validating the answers generated by \textsc{PathFid} are significantly grounded in the supporting facts it generates, which might implicitly indicate that the generated reasoning path tightly aligns with the model's underlying process for answer generation.
Although this is a strong evidence, it is still quite implicit in exposing the model's prediction process, so we see our approach as a step in the right direction rather than a complete solution.

\noindent{\textbf{Performance breakdown by the number of supporting facts and question types}.}
In Table \ref{table:breakdown}, we compare the performance of models by breaking them down based on the number of gold supporting sentences and the question type (e.g., bridge and comparison).
Our first observation is that \textsc{PathFid} provides consistent improvement on answer-EM score over \textsc{Fid} across both the question types and different number of supporting facts required to answer the question.
The high variance in the answer-EM score on \textit{comparison} questions can be attributed to the strictness of exact-match metric as well as the imbalanced nature of the dataset where only 5\% of the comparison questions have more than 3 supporting facts.
Surprisingly, both \textsc{FiD} and \textsc{PathFiD} models perform considerably well on the \textit{comparison} questions even when it requires at least 5 supporting facts. 

A more important motivation behind the performance breakdown analysis was to understand how the supporting fact prediction of \textsc{PathFid} would change as the number of gold supporting facts grows. 
Although it starts degrading on examples with more than 2 supporting facts, it still achieves more than 25\% Support-EM for \textit{bridge} questions with up to 4 supporting facts.
Recalling the average performance on the whole dataset is less than 60\%, we conclude this result might be satisfactory enough, especially for a fully generative model on a very strict evaluation metric.

\noindent{\textbf{Analyzing the evolution of sub-tasks during joint training with \textsc{PathFid}.}} 
\begin{figure}[t]
    \center
	\includegraphics[width=0.48\textwidth]{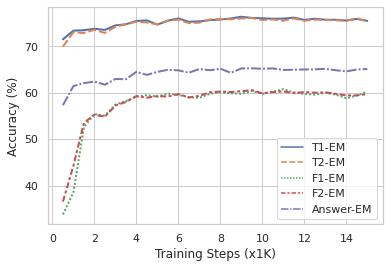}
	\caption{\textsc{PathFid} model evolution on the HotpotQA Dev set during training. T1-EM, T2-EM, indicate the model's accuracy on predicting the title-1 and title-2 on the reasoning path. Similarly F1-EM, and F2-EM denote the model's accuracy on predicting set of supporting facts in passage-1 and passage-2.}
	\label{figure:model_evolution}
% 	\vspace{-2mm}
\end{figure}
In Figure~\ref{figure:model_evolution}, we present the evolution of \textsc{PathFid} model on the HotpotQA development set at every 500 training steps. 
We observe that while the model more quickly picks up the patterns for title generation, it takes much longer for it to reach to a reasonable level of fact prediction.
As one would expect, the general trend in the evolution of different segments (title-1, facts-1, title-2, facts-2, answer) of the reasoning path mostly follows the difficulty of the corresponding sub-task although all the sub-tasks are jointly formulated and trained in an end-to-end fashion.
On the other hand, it seems counter-intuitive for model to reach to a better accuracy on predicting the facts of the second passage (F2-EM) on the reasoning path earlier despite having a better accuracy on (T1-EM).
However, one can also interpret it as a result of stronger feedback provided by the answer segment of the reasoning path as most of the ground-truth answers are contained in the facts of the second passage.

\begin{table*}[t]
	\centering
	\begin{adjustbox}{max width=\textwidth}
		\begin{tabular}{lcccccc}
			\hline  \\ [-2ex]
			\multicolumn{1}{c}{} & \multicolumn{3}{c}{\textbf{Generated-Title EM}} & \multicolumn{3}{c}{\textbf{Reconstructed-Title EM}} \\
			\cline{2-4} \cline{5-7} \\ [-2ex]
			\textbf{Reasoning Path} &  Passage-1 & Passage-2 & Passage-Chain & Passage-1 & Passage-2 & Passage-Chain \\
			\hline
			\hline  \\ [-2ex]
			\textbf{[t1-t2]} & 74.3  & 74.8 & 71.6 & 75.4 & 75.4 & 72.9  \\
			\textbf{[t1-t2-answer]} & 74.8 & 75.0 & 71.8 & 75.8  & \textbf{75.8} & \textbf{73.3}  \\
            \textbf{[t1-f1-t2-f2-answer]} & \textbf{75.0}  & \textbf{75.1} & \textbf{71.9} & \textbf{76.0} & 75.6 & \textbf{73.3} \\
            \hline \\ [-2ex]
		\end{tabular}
	\end{adjustbox}
	\caption[Table caption text]{The effect of joint training as a case study on title prediction performance of \textsc{PathFid} variants trained with different target reasoning paths. \textbf{Generated-Title} column corresponds to ordered passage chain prediction performance in exact-match (EM), while \textbf{Reconstructed-Title} version is computed after applying title reconstruction post-processing described in Section \ref{section:appendix_details}.}
	\label{table:appendix_joint_training_analysis}
\end{table*}

\subsection{Experiments: IIRC}
In addition to our main experiments presented in greater detail, we also conduct experiments on IIRC dataset to verify the generalization of the proposed approach.
To this end, we closely follow the authors' model-free retrieval setting (referred to as Oracle L+C in Table-3) because the model checkpoints for the baseline retrieval model are not available in the public release. 
We use a python script\footnote{\url{https://github.com/jferguson144/IIRC-baseline/blob/main/make_drop_style.py}} provided in the open-sourced repository to replicate the same setting for a fair comparison.
\begin{table}[t]
	\centering
	\begin{adjustbox}{max width=\textwidth}
		\begin{tabular}{lcc}
			\hline  \\ [-2ex]
			\multicolumn{1}{c}{} & \multicolumn{2}{c}{\textbf{Answer}} \\
			\cline{2-3} \\ [-2ex]
			\textbf{Methods} &  EM &  F1 \\
			\hline
			\hline  \\ [-2ex]
            IIRC* \citep{ferguson-2020-iirc} & 63.9 & 69.2  \\
            \textsc{Fid}** \citep{izacard2020fid} & 63.4 & 69.1  \\
			\hline
			\textbf{This Work} \\
			\textsc{PathFid}  & 65.2 & 70.5  \\
            \textsc{PathFid+} & 68.1 & 72.9  \\
			\hline
		\end{tabular}
	\end{adjustbox}
% 	\vspace{1mm}
	\caption[Table caption text]{Experimental results on IIRC dataset in model-free retrieval setting comparing the proposed method against two baselines. * indicates that the result is taken directly from the original paper \citep{ferguson-2020-iirc} (see their Table-3), while ** indicates that we obtain the result of \textsc{Fid} with our implementation.} %\semih{Review the Caption}
	\label{table:iirc_res}
\end{table}

In Table \ref{table:iirc_res}, we present the results on the development set for our proposed \textsc{PathFiD} and \textsc{PathFiD+} in comparison with the baseline reported in the original paper \citep{ferguson-2020-iirc} and our implementation of the FiD~\citep{izacard2020fid} baseline.
\textsc{FiD} model obtains a comparable F1 with IIRC baseline with a slightly worse exact-match performance.
However, the proposed \textsc{PathFid} approach is able to provide 1.3\% and 1.4\% improvement in F1 score over the two baselines.
Furthermore, \textsc{PathFid+} extension leads to the best performance achieving 4.7\% and 4.2\% EM score improvement in absolute value over the \textsc{Fid} baseline and IIRC baseline, respectively.
Our experimental results validate the benefit of the proposed approach on the IIRC dataset, suggesting strong evidence for the generalizability of our approach.

\subsection{Analyzing the Benefit of Joint Training}
In Table \ref{table:appendix_joint_training_analysis}, we present the results of a case study where we analyze the benefit of multi-task training on the passage chain prediction.
The first row of the table shows the results for training \textsc{PathFid} only to predict the sequence of titles for the gold passages (i.e., \textbf{[t1-t2]}), which is just a subsequence of the full reasoning path obtained by discarding facts and the answer. 
The second row is another variant, where we add the answer back to the linearized target sequence while still excluding the segments corresponding to the facts.
The last row correspond to the full reasoning path generation, which is corresponding to the original formulation of \textsc{PathFid} as described in Section \ref{section:approach_pathfid} and illustrated in Figure \ref{figure:model_overview}.
Comparing first two rows in Table \ref{table:appendix_joint_training_analysis}, we can immediately observe that including answer segment in the target reasoning path (i.e., \textbf{[t1-t2-answer]}) boosts the performance across the board although in principle it makes the task more complicated while utilizing the same underlying model capacity.
Further including segments corresponding to \textsc{facts} (sentences within supporting passages) in addition to answer segment (i.e., \textbf{[t1-f1-t2-f2-answer]} -- full reasoning path) boosts the title-EM even further, especially before applying title reconstruction post-processing step. 
Although the objective of the first task (i.e., \textbf{[t1-t2]}) is perfectly aligned with the evaluation metric used in Table \ref{table:appendix_joint_training_analysis}, the performance of the resulting model remains inferior compared to jointly modeling the same task with the answer (and/or supporting facts) prediction.
These two observations elicit a compelling evidence regarding the benefit of jointly modeling the sub-tasks of multi-hop QA as single sequence capturing the full reasoning path.

    \section{Related Work}
\noindent{\textbf{Multi-hop question answering.}}
Research on multi-hop QA aims to tackle complex questions that require reasoning across multiple pieces of evidence in multiple documents~\citep{welbl-etal-2018-constructing,yang-etal-2018-hotpotqa,ferguson-2020-iirc}.
In particular, the HotpotQA dataset~\citep{yang-etal-2018-hotpotqa} provides both the closed and open-domain settings to evaluate multi-hop reading comprehension models.
Compared to single-hop QA, such complex questions pose additional challenges for both reader and retriever models since they are required to capture relationships between documents, instead of independently processing each document.
This is challenging because the number of document combinations exponentially grows due to the sequential nature of the process.
Two recent works~\citep{nie-etal-2019-revealing,asai2020grr} have tackled this challenge by leveraging hyperlink structure in the underlying Wikipedia corpus, while \citet{xiong2021mdr} has taken a recursive approach to extend the dense retrieval process to handle sequential search.
Most of the reading comprehension (RC) models in existing work~\citep{xiong-etal-2019-simple,chen2019bertpara,nishida-etal-2019-answering,qi2021retrieve,li2020hopretriever,xiong2021mdr} follow an extractive architecture~\citep{devlin-etal-2019-bert} for selection of the answer spans and their corresponding supporting evidence with minor modifications such as initializing the backbone model from a stronger or larger pre-trained models \citep{clark2020electra}.
On the other hand, some recent works~\citep{inoue-etal-2021-summarize} take a more abstractive approach and generate question-focused summaries of input paragraphs as concise explanations to be fed to the RC module.

\noindent{\textbf{Generative question answering.}}
Especially after the emergence of the SQuAD dataset~\citep{rajpurkar-etal-2016-squad}, neural extractive QA models have been widely studied.
An underlying assumption is that we can extract a short text span (or a phrase) as an answer, but it is not always the case in reality.
Motivated by this, the generative QA approach has also been investigated~\citep{hewlett-etal-2017-accurate,fan2019eli5}.
Recent advances on pre-trained transformers have pushed this direction; for example, \citet{lewis-etal-2020-bart} jointly trained a generative QA model along with a text retrieval model, and \citet{roberts-etal-2020-much} explored an ambitious approach to directly generate an answer without any evidence documents.
We focused on the fusion-in-decoder model~\citep{izacard2020fid}; they claimed that the decoder might be good at aggregating information across multiple documents.
However, we have shown that it is not trivial in the multi-hop reasoning task, and pushed the model's ability to jointly learn to predict reasoning paths.
Besides question answering, jointly learning multiple intrinsic capabilities required by the final objective with a generative approach has been shown useful in modeling other NLP tasks such as task-oriented dialogues~\citep{neelakantan2019assistant,hosseini2020simpletod,peng-etal-2021-soloist}.

\noindent{\textbf{Open-domain question answering.}}
Open-domain QA~\citep{voorhees1999trec8} is practically important, which requires a system to retrieve relevant documents to answer a given question.
The task is recently gaining much attention, thanks to the development of large-scale datasets like HotpotQA, SQuAD Open~\citep{chen-etal-2017-reading}, Natural Questions Open~\citep{kwiatkowski-etal-2019-natural,lee-etal-2019-latent}, etc.
Pre-trained transformer models like BERT~\citep{devlin-etal-2019-bert} have accelerated the development of neural text retrievers~\citep{lee-etal-2019-latent,karpukhin-etal-2020-dense,asai2020grr,xiong2021mdr,liu-2021-dhr} in the retriever-reader framework~\citep{chen-etal-2017-reading}.
We have investigated the effectiveness of our method in the multi-hop open-domain QA task (see Appendix \ref{appendix:full_wiki}) using an existing external retriever component.

    \section{Conclusion}

In this work, we propose a generative question answering (QA) approach that models multi-hop QA as a single sequence prediction task.
It learns to generate an answer along with a reasoning path to improve its capability of multi-hop reasoning. 
Our experiments on prominent multi-hop QA benchmarks, HotpotQA and IIRC, validate the promise and effectiveness of our proposed method \textsc{PathFid} and its extension \textsc{PathFid+}.
Future work will explore (1) our \textsc{PathFid} approach more closely with text retrieval models in open-domain QA scenarios and (2) more explicit grounding on the input information to make our approach even more interpretable and controllable.

    \section*{Acknowledgments}

The authors would like to thank the members of Salesforce AI Research team for fruitful discussions, as well as the anonymous
reviewers for their helpful feedback.

    \bibliography{references}
    \bibliographystyle{acl_natbib}
    
    \appendix
\clearpage
\section{Visualizing the Correlation between Evidence and Answer}
\begin{figure}[h]
    \center
	\includegraphics[width=0.5\textwidth]{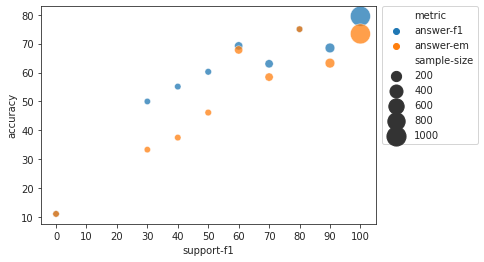}
	\caption{Visualizing the correlation between evidence and answer prediction for COMPARISON questions.}
	\label{figure:appendix_plot_comparison_correlation}
\end{figure}

\begin{figure}[h]
    \center
	\includegraphics[width=0.5\textwidth]{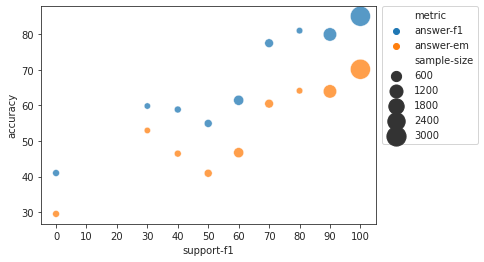}
	\caption{Visualizing the correlation between evidence and answer prediction for BRIDGE questions.}
	\label{figure:appendix_plot_bridge_correlation}
\end{figure}

In Figure \ref{figure:appendix_plot_comparison_correlation} and \ref{figure:appendix_plot_bridge_correlation}, we visualize the correlation between supporting evidence and answer prediction performances for \textit{comparison} and \textit{bridge} question types, respectively.
To obtain these plots, we first split the examples into 10 buckets where $n$-th bucket contains the examples with support-F1 score in $(10*(n-1), 10*n]$ percentile for $n=\{1, 2, \ldots, 10\}$. 
Then, we take the average answer prediction accuracy (both EM and F1) over these examples for each bucket, and report this number on the y-axis of the plot at the corresponding support-F1 bucket on the x-axis, while dropping the empty buckets.
Note that $x=0$ corresponds to examples with support-F1 score of 0.
Also note that the size of a data point on the figure reflects the number of examples in the corresponding bucket as also indicated by the legend.
From Figures \ref{figure:appendix_plot_comparison_correlation} and \ref{figure:appendix_plot_bridge_correlation}, we can observe that the accuracy of the generated answers is significantly lower, ~30\% for \textit{bridge} and ~10\% for \textit{comparison}, for the first bucket with zero support-F1 compared to buckets with positive support-F1 score.
This suggests that the model has a difficult time figuring out the answer when the supporting evidence prediction is poor.
Another observation that holds for both categories is the general trend of increased answer quality as the supporting fact prediction improves. 
Combining these two points provide additional evidence (in addition to Table \ref{table:groundedness} in the main paper) implicitly supporting the answer generation process of \textsc{PathFid} being grounded on the generated supporting facts, which is generated as the prefix of the answer segment in the full decoded reasoning path sequence during inference.

\section{Case Study: Full-Wiki Setting with Multi-hop Dense Retriever} \label{appendix:full_wiki}
\begin{table*}[!t]
	\centering
	\begin{adjustbox}{max width=\textwidth}
		\begin{tabular}{lcccc}
			\hline  \\ [-2ex]
			\multicolumn{1}{c}{} & \multicolumn{2}{c}{\textbf{Answer}} & \multicolumn{2}{c}{\textbf{Support}} \\
			\cline{2-5} \\ [-2ex]
			\textbf{Methods} &  EM &  F1 & EM &  F1 \\
			\hline
			\hline  \\ [-2ex]
            GoldEn Retriever \ \ \citep{qi-etal-2019-answering}    &  - & 49.8 & - & 64.6  \\
            Semantic Retrieval \ \ \citep{nie-etal-2019-revealing} &  46.5 & 58.8 & 39.9 & 71.5 \\
            Transformer-XH \ \ \citep{zhao2020transformerxh}       & 50.2 & 62.4 & 42.2 & 71.6 \\
            Graph Recurrent Retriever \citep{asai2020grr} (wwm)    & 60.5 & 73.3 & 49.3 & 76.1  \\
            Graph Recurrent Retriever \citep{asai2020grr} (base)   & 52.7 & 65.8 & 47.9 & 75.0  \\
            HopRetriever \citep{li2020hopretriever}                & 62.1 & 75.2 & 52.5 & 78.9  \\
            HopRetriever-plus \citep{li2020hopretriever}           & 66.6 & 79.2 & 56.0 & 81.8  \\
			\hline
			MDR-Electra (Top-50 paths) \citep{xiong2021mdr} &  61.7 & 74.3 & - & - \\
			MDR-FiD (Top-50 paths) \citep{xiong2021mdr} &  61.7 & 73.1 & - & - \\
			\hline
			\textbf{Our Models} \\
			\textsc{Fid}* (Top-25 paths) & 54.0 & 66.0 & - & -  \\
			\textsc{PathFid} (Top-25 paths)  & 55.8 & 67.9 & 49.0 & 74.1  \\
			\textsc{PathFid+} (Top-25 paths)  & 59.8 & 72.4 & 52.8 & 76.6  \\
			\hline
			\textbf{On} $\textbf{Dev}^{*}$ \textbf{Evaluation} \\
			\textsc{PathFid+} (Top-25 paths)  & 70.2 & 81.5 & 60.9 & 86.3  \\
			\hline
		\end{tabular}
	\end{adjustbox}
	\vspace{1mm}
	\caption[Table caption text]{Results for open-domain setting using MDR \citep{xiong2021mdr} as the retriever. $\text{Dev}^{*}$ refers to the development set where the retrieved passages are expanded with the gold passage (as an oracle setting) to account for the cases where the retriever fails to retrieve the gold passages. \textsc{Fid}* indicates our implementation.}
	\label{table:open_setting}
\end{table*}
In this subsection, we evaluate \textsc{PathFid} in open domain setting of HotpotQA leveraging a recently proposed multi-hop dense retriever (MDR) \citep{xiong2021mdr} for passage retrieval.
Unlike distractor setting, MDR returns a set of passage pairs $D_q^{\text{MDR}} = \{(p_1^{(1)},p_1^{(2)}), (p_2^{(1)},p_2^{(2)}), \ldots, (p_N^{(1)},p_N^{(2)})\}$ for question $q$, where each passage $p_n^{(i)}$ comes with a title $t_n^{(i)}$, being retrieved from Wikipedia corpus.
This setting naturally fits into how we formulate \textsc{PathFid+}, which operates on the pairs of input passages as introduced in Section \ref{subsection:path_fid_plus}, where we simply set $D_q^{+} = D_q^{\text{MDR}}$.
For experiments with \textsc{Fid} and \textsc{PathFid}, which operate on set of single input passages, we simply split the pairs into single passages, ending up with $2K$ passages when using top-$K$ retrieved paths from MDR.
We present our results for this setting in Table \ref{table:open_setting}. Similar to our observation in distractor setting, \textsc{PathFid} provides a significant (\%1.8) answer EM score improvement over \textsc{Fid}, while also achieving a quite competitive performance on the supporting fact prediction compared to strong discriminative models \citep{asai2020grr, li2020hopretriever} optimized for better retrieval performance.
Most notably, \textsc{PathFid+} provides significant gains over \textsc{PathFid}, achieving 59.8\% answer-EM and 52.8\% supporting fact EM score, showing the importance of encoding cross-passage interactions.
It is important to note here that our results with \textsc{PathFid+} is not directly comparable to the reader results from MDR \citep{xiong2021mdr} because we are able to only use top-25 retrieved paths due to hardware limitations.
Finally, we also evaluate the same \textsc{PathFid+} model on $\text{Dev}^{*}$ obtained by adding the pair of gold passages in $D_q^{\text{MDR}}$, where we aim to isolate the error propagation from the underlying retriever.
Table \ref{table:open_setting} shows that both the answer and supporting fact prediction performance improves quite significantly, showing the potential impact that developments on retriever side of the problem can also make.

\section{The Effect of Model Size for Future Reference}
\begin{table}[!t]
	\centering
	\small
	\begin{adjustbox}{max width=\textwidth}
		\begin{tabular}{llcccc}
			\hline  \\ [-2ex]
			\multicolumn{1}{c}{} & \multicolumn{1}{c}{} & \multicolumn{2}{c}{\textbf{Answer}} & \multicolumn{2}{c}{\textbf{Support}} \\
			\cline{3-6} \\ [-2ex]
			\textbf{Model Size} & \textbf{Top-K Paths} &  EM &  F1 & EM &  F1 \\
			\hline \\ [-2ex]
			\textsc{T5-base}  & Top-25 &  56.6 & 69.1 & 51.9 & 75.7  \\
			\hline  \\ [-2ex]
            \textsc{T5-large}  & Top-25 &  59.8 & 72.4 & 52.8 & 76.6  \\
			\hline
		\end{tabular}
	\end{adjustbox}
	\vspace{1mm}
	\caption[Table caption text]{Full-wiki results with \textsc{PathFid+} comparing two different T5 model sizes.}
	\label{table:appendix_model_size}
\end{table}
As discussed in Section \ref{section:appendix_details}, fine-tuning \textsc{PathFid+} with T5-large initialization might require significant resources and non-trivial memory efficient optimization (e.g., gradient checkpointing).
To provide a baseline with a smaller model for future research, here we include the results of \textsc{PathFid+} with T5-base initialization using the same setting reported in Table \ref{table:open_setting} in the main paper.
As presented in Table \ref{table:appendix_model_size}, although the performance difference on the supporting fact prediction is relatively small (~1\%), answer prediction performance drops significantly (by 3.2\%) when we switch from T5-large to T5-base.
However, working with T5-base is much more efficient in terms of resources and iteration time for building baselines, trying out new ideas and thought experiments.
So, we hope this baseline will be helpful for future research.

\section{More on Training and Implementation Details} \label{section:appendix_details}
\noindent{\textbf{Hop ordering.}} HotpotQA benchmark provides annotation only for \textit{unordered} gold passages, without explicitly specifying which passage corresponds to the $k$-th hop (e.g., first-hop, second-hop, etc.) on the reasoning path. 
In our implementation, we combine the heuristic strategies applied by GRR \citep{asai2020grr} and MDR \citep{xiong2021mdr}. 
More precisely, if only one of the gold passages contains the answer, then we take the passage that includes the answer span as the final passage. If the answer span is included in both passages, we break the tie by falling back to the hyperlink-based ordering strategy proposed by GRR \citep{asai2020grr}.

\noindent{\textbf{Post-processing for passage title reconstruction.}}
Note that \textsc{PathFid} generates the titles of the passages on the reasoning path token by token including the separator tokens. 
However, the decoder might fall into some minor errors during the generation process, which may cause the resulting titles to end up slightly different from the original ones. 
To account for such minor errors, we leverage the set of titles coming from the input passages and find the most similar among them to our generated passage titles based on token-level F1-score. 
We call this process \textit{title reconstruction} and apply it while reporting the performance for supporting fact predictions. 
Table \ref{table:appendix_joint_training_analysis} shows the benefit of \textit{title reconstruction} for mitigating such minor generation errors. 
On the other hand, the small performance boost suggests that titles \textsc{PathFid} already generates quite faithful title predictions.

\noindent{\textbf{Model selection.}} For all the models reported in this work, we perform evaluation at every 500 steps during training by decoding the whole development set on a separate machine in a non-blocking fashion. 
We then select the best model based on the answer exact-match score performance.
However, since \textsc{PathFid} variants generate more than just the answer, it can be leveraged to optimize for a more holistic metric including the supporting fact prediction performance, offering further control on model selection.
We leave further exploration of this phenomenon to future work.

\noindent{\textbf{Scaling to larger evidence pools for full-wiki setting.}}
As briefly noted in Appendix \ref{appendix:full_wiki}, we report results in full-wiki setting using only top-25 paths returned by MDR \citep{xiong2021mdr} due to hardware constraints. 
More precisely, a single training example becomes impossible to fit into GPU memory (40GB) even for top-25 paths for \textsc{PathFid+} model with T5-large initialization.
To make the training feasible, we resort to gradient checkpointing\footnote{https://pytorch.org/docs/stable/checkpoint.html} which trades off GPU memory with speed. 
However, in this case, even with 25 retrieved paths, training \textsc{PathFid+} for 10K steps with batch size of 64 using gradient accumulation takes ~19 hours on 8 A100 GPUs with 40GB memory each, which is one of the most prominent limitations hurdling the progress for this line of research.
Further research on making generative approaches with large pre-trained models more efficient without losing on the performance side holds a great potential impact to accelerate the progress of fully generative models for question answering.

\section{Hyperparameter Settings}
\begin{table}[!t]
	\centering
	\scriptsize
	\begin{adjustbox}{max width=\textwidth}
		\begin{tabular}{lccc}
		    \hline  \\ [-2ex]
		    parameter & \textsc{FiD} & \textsc{PathFid} & \textsc{PathFid+} \\
			\hline  \\ [-2ex]
			initialization &  t5-large &  t5-large  &  t5-large \\
            learning rate &  1e-4  &  1e-4  &  1e-4 \\
            learning rate schedule & constant & constant & constant  \\
            batch size &  64 &  64 &  64 \\
            gradient checkpointing &  no &  no &  no \\
            maximum input length   & 256 & 256 & 512  \\
            maximum output length  & 32 & 64 & 64  \\
            warmup ratio           & 0 & 0  & 0 \\
            gradient clipping norm & 1.0 & 1.0 & 1.0  \\
            training epoch         & 10 & 10 & 10 \\
            weight decay           & 0 & 0 & 0 \\
			\hline
		\end{tabular}
	\end{adjustbox}
	\vspace{1mm}
	\caption[Table caption text]{Hyperparameters for experiments on HotpotQA Distractor setting.}
	\label{table:appendix_hparams_distractor}
\end{table}

\begin{table}[!t]
	\centering
	\scriptsize
	\begin{adjustbox}{max width=\textwidth}
		\begin{tabular}{lccc}
		    \hline  \\ [-2ex]
		    parameter & \textsc{FiD} & \textsc{PathFid} & \textsc{PathFid+} \\
			\hline  \\ [-2ex]
			initialization &  t5-large &  t5-large  &  t5-large \\
            learning rate &  1e-4  &  1e-4  &  1e-4 \\
            learning rate schedule & constant & constant & constant  \\
            batch size &  64 &  64 &  64 \\
            gradient checkpointing &  no &  no &  no \\
            maximum input length   & 256 & 256 & 512  \\
            maximum output length  & 32 & 64 & 64  \\
            warmup ratio           & 0 & 0  & 0 \\
            gradient clipping norm & 1.0 & 1.0 & 1.0  \\
            training epoch         & 10 & 10 & 10 \\
            weight decay           & 0 & 0 & 0 \\
			\hline
		\end{tabular}
	\end{adjustbox}
	\vspace{1mm}
	\caption[Table caption text]{Hyperparameters for experiments on IIRC dataset.}
	\label{table:appendix_hparams_iirc}
\end{table}

\begin{table}[!t]
	\centering
	\scriptsize
	\begin{adjustbox}{max width=\textwidth}
		\begin{tabular}{lccc}
		    \hline  \\ [-2ex]
		    parameter & \textsc{FiD} & \textsc{PathFid} & \textsc{PathFid+} \\
			\hline  \\ [-2ex]
			initialization &  t5-large &  t5-large  &  t5-large \\
            learning rate &  1e-4  &  1e-4  &  1e-4 \\
            learning rate schedule & constant & constant & constant  \\
            batch size &  64 &  64 &  64 \\
            gradient checkpointing &  yes &  yes &  yes \\
            maximum input length   & 256 & 256 & 512  \\
            maximum output length  & 32 & 64 & 64  \\
            warmup ratio           & 0 & 0  & 0 \\
            gradient clipping norm & 1.0 & 1.0 & 1.0  \\
            training steps         & 10K & 10K & 10K \\
            weight decay           & 0 & 0 & 0 \\
            top-K path retrieval   & 25 & 25 & 25 \\
			\hline
		\end{tabular}
	\end{adjustbox}
	\vspace{1mm}
	\caption[Table caption text]{Hyperparameters for experiments on HotpotQA Full-wiki setting.}
	\label{table:appendix_hparams_open}
\end{table}
In Tables \ref{table:appendix_hparams_iirc}, \ref{table:appendix_hparams_distractor} and \ref{table:appendix_hparams_open}, we provide the full set of important hyperparameters used for the models reported both in the main paper (HotpotQA-distractor and IIRC) and in the Appendix \ref{appendix:full_wiki} (HotpotQA-fullwiki), respectively.

\section{Qualitative Analysis} \label{appendix:qualitative_analysis}
\begin{table*}[hbt!]
    \centering
    \footnotesize
    \begin{tabular}{l|p{10cm}}
    % \hline
    \specialrule{.3em}{.2em}{.2em}

    \multirow{1}{*}{\textbf{Question}} & The Memphis Hustle are based in a suburb of a city with a population of what in 2010?
\\
        \hline
        \multirow{13}{*}{\textbf{Input Passages}} 
        & {\underline{\textcolor{blue}{1. Memphis Hustle:}} <f1> \textcolor{orange}{The Memphis Hustle are an American professional basketball team of the NBA G League announced to begin play for the 2017–18 season as an affiliate of the Memphis Grizzlies of the National Basketball Association (NBA)}. <f2> \textcolor{orange}{Based in the Memphis suburb of Southaven, Mississippi, the team will play their home games at the Landers Center}.} \\
        & {\underline{\textcolor{blue}{2. Southaven, Mississippi:}} <f1> \textcolor{orange}{Southaven is a city in DeSoto County, Mississippi, United States}. <f2> It is a suburb of Memphis, Tennessee, and a principal city in the Memphis metropolitan area. <f3> \textcolor{orange}{The 2010 census reported a population of \underline{\textcolor{green}{48,982}}, making Southaven the third largest city in Mississippi}. <f4> Southaven is traversed from north to south by the I-55/I-69 freeway. <f5> The city's name derives from the fact that Southaven is located south of Whitehaven, a neighborhood in Memphis.} \\
        & {\underline{\textcolor{blue}{3. Lakeland, Tennessee:}} Lakeland is a city in Shelby County, Tennessee, and a suburb of Memphis. \textcolor{red}{The population was \underline{12,430} at the 2010 census.}} \\
        & {\underline{\textcolor{blue}{4. Marion, Arkansas:}} Marion is a city in and the county seat of Crittenden County, Arkansas ...} \\
        & {\underline{\textcolor{blue}{5. West Memphis, Arkansas:}} West Memphis is the largest city in Crittenden County, Arkansas ...} \\
        & ...
 \\
        \hline \\ [-2ex]
        \multirow{1}{*}{\textbf{Gold Answer}}  & \textbf{48,982}
 \\
%         \hline
%         \multirow{1}{*}{\textbf{Target RP}}  & \textcolor{blue}{<|title-1|>} 
%         asdadsa \textcolor{blue}{<|title-2|>} asdadsad
%  \\
        \hline
        \multirow{1}{*}{\textbf{\textsc{Fid} Answer}} & \textcolor{red}{12,430}
\\
        \hline
        \multirow{1}{*}{\textbf{\textsc{PathFid} Answer}} & \textcolor{green}{48,982}
\\
        \hline
        \multirow{2}{*}{\textbf{\textsc{PathFid} Output}} & \textcolor{blue}{<title-1>} Memphis Hustle \textcolor{orange}{<facts-1>} <f1> <f2> \textcolor{blue}{<title-2>} Southaven, Mississippi \textcolor{orange}{<facts-2>} <f1> <f2> <f3> \textcolor{green}{<answer>} 48,982
\\         
    %  \hline
    %  \hline
    \specialrule{.3em}{.2em}{.2em}

    \end{tabular}
    \vspace{1.5mm}
    \caption{BRIDGE-type question example, where \textbf{\textsc{PathFid} predicts the correct answer while \textsc{Fid} fails to do so.} The third passage is the \textbf{distractor causing \textsc{Fid} to make a wrong prediction} due to the highlighted sentence in \textcolor{red}{red}.}
    \label{table:example_bridge1}
\end{table*}
In this section, we provide examples comparing the predictions of \textsc{Fid} and \textsc{PathFid} over \textit{bridge} and \textit{comparison} question types.
Each of the example Table \ref{table:example_bridge1}, \ref{table:example_bridge2}, \ref{table:example_comp} in the next pages follows a similar structure, where we include gold answer, \textsc{Fid} answer prediction, \textsc{PathFid} answer (and full path) prediction, and 5 supporting passages (out of 10) for the brevity of presentation.
Among the input passages, the first two correspond to gold passages, for which we include the full content as well as highlighting the key supporting facts/sentences with orange color.
The following three passages are presented as a subset of the distractors, for each of which we include a one-line content unless it plays a crucial role in distracting at least one of the models in making a wrong prediction.
In this case, we also add the content of this particular passage as well as highlighting the specific distractor span/sentence causing the failure of either \textsc{Fid} or \textsc{PathFid}.
\clearpage
%%%%%%%%%%%%%%% BRDIGE both WRONG %%%%%%%%%%%%%%%%%%
\begin{table*}[ht]
    \centering
    % \scriptsize
    \footnotesize
    \begin{tabular}{l|p{10cm}}
    % \hline
    \specialrule{.3em}{.2em}{.2em}

    \multirow{2}{*}{\textbf{Question}} & What government position was held by the woman who portrayed Corliss Archer in the film Kiss and Tell?
\\
        \hline
        \multirow{13}{*}{\textbf{Input Passages}} 
        & {\underline{\textcolor{blue}{1. Kiss and Tell (1945 film):}} <f1> \textcolor{orange}{Kiss and Tell is a 1945 American comedy film starring then 17-year-old Shirley Temple as Corliss Archer}. <f2> In the film, two teenage girls cause their respective parents much concern when they start to become interested in boys. <f3> The parents' bickering about which girl is the worse influence causes more problems than it solves.} \\
        & {\underline{\textcolor{blue}{2. Shirley Temple:}} <f1> \textcolor{orange}{Shirley Temple Black (April 23, 1928 – February 10, 2014) was an American actress, singer, dancer, businesswoman, and diplomat who was Hollywood's number one box-office draw as a child actress from 1935 to 1938}. <f2> \textcolor{orange}{As an adult, she was named \underline{\textcolor{red}{United States ambassador}} to Ghana and to Czechoslovakia and also served as \underline{\textcolor{green}{Chief of Protocol of the United States}}}.} \\
        & {\underline{\textcolor{blue}{3. Meet Corliss Archer (TV series):}} Meet Corliss Archer is an American television sitcom that ...} \\
        & {\underline{\textcolor{blue}{4. Meet Corliss Archer:}} Meet Corliss Archer, a program from radio's Golden Age, ran from ...} \\
        & {\underline{\textcolor{blue}{5. Charles Craft:}} Charles Craft (May 9, 1902 – September 19, 1968) was an English-born ...} \\
        & ...
 \\
        \hline \\ [-2ex]
        \multirow{1}{*}{\textbf{Gold Answer}}  & \textbf{Chief of Protocol}
 \\
%         \hline
%         \multirow{1}{*}{\textbf{Target RP}}  & \textcolor{blue}{<|title-1|>} 
%         asdadsa \textcolor{blue}{<|title-2|>} asdadsad
%  \\
        \hline
        \multirow{1}{*}{\textbf{\textsc{Fid} Answer}} & \textcolor{red}{United States ambassador}
\\
        \hline
        \multirow{1}{*}{\textbf{\textsc{PathFid} Answer}} & \textcolor{green}{Chief of Protocol of the United States}
\\
        \hline
        \multirow{2}{*}{\textbf{\textsc{PathFid} Output}} & \textcolor{blue}{<title-1>} Kiss and Tell (1945 film) \textcolor{orange}{<facts-1>} <f1> \textcolor{blue}{<title-2>} Shirley Temple \textcolor{orange}{<facts-2>} <f2> \textcolor{green}{<answer>} Chief of Protocol of the United States
\\         
    %  \hline
    %  \hline
    \specialrule{.3em}{.2em}{.2em}

    \end{tabular}
    \vspace{1.5mm}
    \caption{BRIDGE-type question example, where both \textbf{\textsc{PathFid} and \textsc{Fid} fail to predict the exact gold answer.} Although the generated answers are wrong, they can both be acceptable by humans. On the other hand, both answers fail in EM accuracy, but \textsc{PathFid} manages to perfectly generate the reasoning path starting from the right sentence of the correct first passage, then jumping to correct second-hop passage, followed by identifying its key sentence (<f2>), then finally locating answer in the right part of this evidence, but only failing in getting the span perfectly, which still rewards it with a reasonable F1 score. However, this example is also important in showing the possible ambiguities in questions and strictness of the exact-match accuracy metric.}
    \label{table:example_bridge2}
\end{table*}
% \begin{table}[htb!]
\begin{table*}[ht]
    \centering
    % \scriptsize
    \footnotesize
    \begin{tabular}{l|p{10cm}}
    % \hline
    \specialrule{.3em}{.2em}{.2em}

    \multirow{1}{*}{\textbf{Question}} & Which band, Letters to Cleo or Screaming Trees, had more members?
\\
        \hline
        \multirow{13}{*}{\textbf{Input Passages}} 
        & {\underline{\textcolor{blue}{1. Screaming Trees:}} <f1> \textcolor{orange}{Screaming Trees was an American rock band formed in Ellensburg, Washington in 1985 by vocalist Mark Lanegan, guitarist Gary Lee Conner, bass player Van Conner and drummer Mark Pickerel}. <f2> \textcolor{orange}{Pickerel had been replaced by Barrett Martin by the time the band reached its most successful period}. <f3> Although widely associated with grunge, the band's sound incorporated hard rock and psychedelic elements. <f4> During Screaming Trees' existence the band released seven studio albums, five EPs, and three compilations.} \\
        & {\underline{\textcolor{blue}{2. Letters to Cleo:}} <f1> \textcolor{orange}{Letters to Cleo are an alternative rock band from Boston, Massachusetts, best known for the 1994 single, "Here \& Now", from their full-length debut album, "Aurora Gory Alice"}. <f2> \textcolor{orange}{The band's members are Kay Hanley, Greg McKenna, Michael Eisenstein, Stacy Jones, Scott Riebling, and later, Tom Polce}.} \\
        & {\underline{\textcolor{blue}{3. Change Has Come:}} Change Has Come was the only recording the Screaming Trees released ...} \\
        & {\underline{\textcolor{blue}{4. Jamboree (Beat Happening album):}} Jamboree is the second album by Beat Happening, released ...} \\
        & {\underline{\textcolor{blue}{5. Gary Lee Conner:}} Gary Lee Conner (born Lee Gary Conner on August 22, 1962 in Fort Irwin ...} \\
        & ...
 \\
        \hline \\ [-2ex]
        \multirow{1}{*}{\textbf{Gold Answer}}  & \textbf{Letters to Cleo}
 \\
%         \hline
%         \multirow{1}{*}{\textbf{Target RP}}  & \textcolor{blue}{<|title-1|>} 
%         asdadsa \textcolor{blue}{<|title-2|>} asdadsad
%  \\
        \hline
        \multirow{1}{*}{\textbf{\textsc{Fid} Answer}} & \textcolor{red}{Screaming Trees}
\\
        \hline
        \multirow{1}{*}{\textbf{\textsc{PathFid} Answer}} & \textcolor{green}{Letters to Cleo}
\\
        \hline
        \multirow{2}{*}{\textbf{\textsc{PathFid} Output}} & \textcolor{blue}{<title-1>} Screaming Trees \textcolor{orange}{<facts-1>} <f1> \textcolor{blue}{<title-2>} Letters to Cleo \textcolor{orange}{<facts-2>} <f1> <f2>  \textcolor{green}{<answer>} Letters to Cleo
\\         
    %  \hline
    %  \hline
    \specialrule{.3em}{.2em}{.2em}

    \end{tabular}
    \vspace{1.5mm}
    \caption{COMPARISON-type question example, where \textsc{PathFid} predicts the correct answer while \textsc{Fid} fails to make a correct prediction.}
    \label{table:example_comp}
\end{table*}

\end{document}